\documentclass[conference]{IEEEtran}
\IEEEoverridecommandlockouts
\usepackage[pdftex]{graphicx} 
\usepackage{amsmath} 
\usepackage{float}
\usepackage[labelformat=simple]{subcaption}
\DeclareCaptionLabelSeparator{periodspace}{.\quad}
\usepackage{multirow}
\usepackage{makecell}
\usepackage{todonotes}
\usepackage{balance}

\graphicspath{{./figures/}}

\begin{document}

\title{
Enhancing Safety for Students with Mobile Air Filtration during School Reopening from COVID-19
} 



\author{Haoguang Yang, Mythra V. Balakuntala, Abigayle E. Moser, Jhon J. Quiñones, Ali Doosttalab,\\
Antonio Esquivel-Puentes, Tanya Purwar, Richard M. Voyles, Luciano Castillo, and Nina Mahmoudian
\thanks{H. Yang, M. V. Balakuntala, J. J. Quiñones, and R. M. Voyles (\texttt{\{yang1510, mbalakun, quinone3, rvoyles\}@purdue.edu}) are with Polytechnic Institute, Purdue University.}
\thanks{A. E. Moser (\texttt{aemoser@iastate.edu}{moser34@purdue.edu}) is with Dept. of Aerospace Engineering, Iowa State University, and Dept. of Mechanical Engineering, Purdue Uiversity.}
\thanks{A. Doosttalab, A. Esquivel-Puentes, T. Purwar, L. Castillo, and N. Mahmoudian (\texttt{\{adoostta, hesquiv, tpurwar, castil63, ninam\}@purdue.edu}) are with Dept. of Mechanical Engineering, Purdue University.}
\thanks{This manuscript has been accepted by 2021 IEEE International Conference on Robotics and Automation (ICRA).}
}

\maketitle

\begin{abstract}
The paper discusses how robots enable occupant-safe continuous protection for students when schools reopen. Conventionally, fixed air filters are not used as a key pandemic prevention method for public indoor spaces because they are unable to trap the airborne pathogens in time in the entire room. However, by combining the mobility of a robot with air filtration, the efficacy of cleaning up the air around multiple people is largely increased. A disinfection co-robot prototype is thus developed to provide continuous and occupant-friendly protection to people gathering indoors, specifically for students in a classroom scenario. In a static classroom with students sitting in a grid pattern, the mobile robot is able to serve up to 14 students per cycle while reducing the worst-case pathogen dosage by 20\% and with higher robustness, compared to a static filter. The extent of robot protection is optimized by tuning the passing distance and speed, such that a robot is able to serve more people given a threshold of worst-case dosage a person can receive.
\end{abstract}

\section{Introduction}

Robots were initially proposed as a protective measure for healthcare workers during the the Ebola outbreak in 2014 \cite{Kraft2016RobotForInfectiousDiseases, robotEbola2014Assessment} -- by performing risky but low-skill tasks such as cleaning.
During the COVID-19 pandemic, the urban infrastructure of the hot-spots enable wider and more effective use of robots -- from aerial to ground robots -- to enhance people's safety and reduce the spread \cite{2020RobotsForInfectiousDiseases}. Among them, clinical care robots emerged mostly during the \textit{onset phase} of COVID-19, as in prior pandemics. They were superceded by robots for public safety as the pandemic spreads and policies evolve into \textit{quarantine phase}. Finally, as we enter the \textit{reopening phase}, robots for workplaces and schools become significant. 

\begin{figure}
    \begin{subfigure}{0.7\linewidth}
    \centering
    \includegraphics[width=1.0\textwidth, trim={0 10mm 0 4mm}]{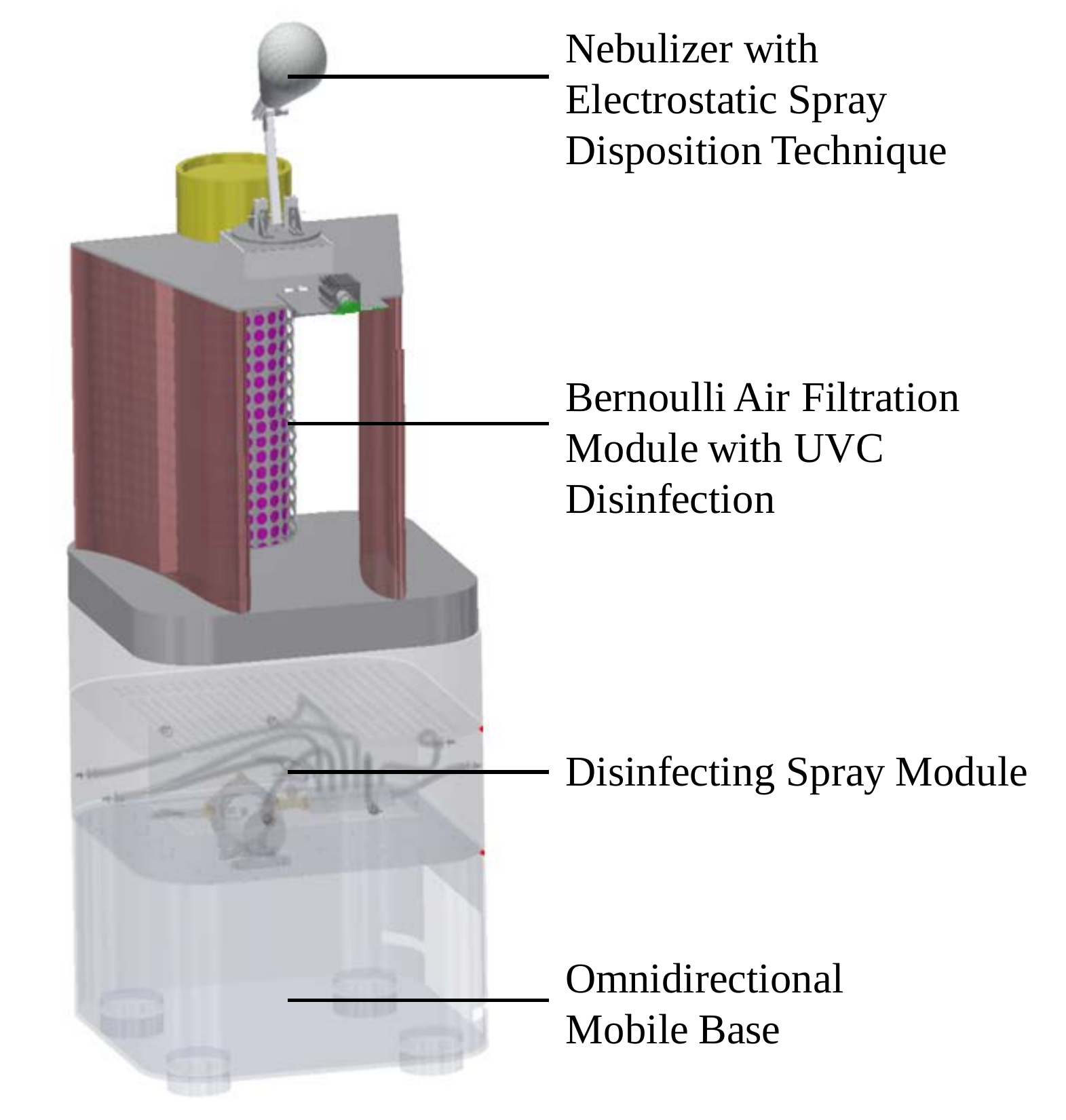}
    \caption{}
    \end{subfigure}
    \begin{subfigure}{0.28\linewidth}
    \centering
    \includegraphics[width=1.0\textwidth, trim={75mm 0 0 0}, clip]{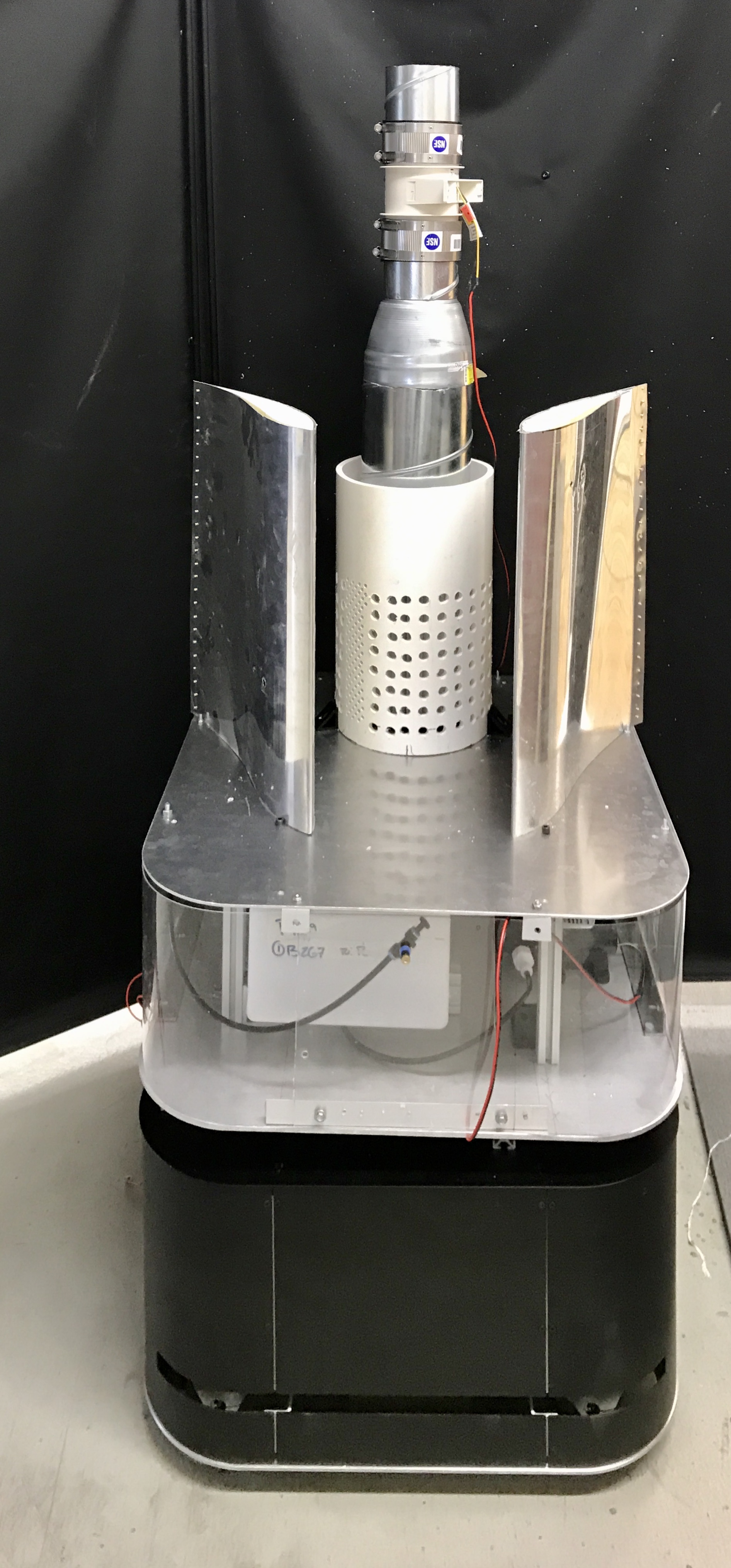}
    \caption{}
    \end{subfigure}
    \caption{The Purdue Campus Patrol Robot (PCPR). (a) The overall assembly and effective disinfection payloads; (b) A physical figure of the robot showing the setup of the Bernoulli Air Filtration Module.}
    \label{fig:robot}
\end{figure}

As PPE supplies catches up and eases pressure on medical staff, community spread becomes the focus of suppressing and reopening from the pandemic. While the exposure rate to virus is low in non-hospital facilities, the risk is accumulated by the long exposure time and number of people involved. Therefore, the necessity lies in preparing educators and essential workers to safely move within suddenly hazardous environments. Driven by these requirements, the Purdue Campus Patrol Robot (PCPR) is developed (Fig. \ref{fig:robot}) to protect people in indoor public places using multimodal disinfection. Among which, the occupant-friendly robotic air filtration fills an urgent need as schools reopen and gatherings in classrooms pose dangerous spreading occasions of the virus through bioaerosol created when sneezing, coughing, or even breathing.

Previous literature featuring robot for disinfection involves either chemical spraying \cite{s20123543SprayRobot, 6840192WallCleaningRobot}, or ultraviolet (UV) light exposure \cite{YANG2019487UVC, RobotInContagiousWorld}. While these methods kills the pathogens directly and are also involved in our robot design, they poses irritation to people around it, hence they are not suitable for full-time protection of indoor personnel. Conventionally, stationary air filtration systems alone is not effective for epidemic prevention \cite{Lee_2011_BioaerosolControl}, because the pathogen may find its victim before being circulated to and trapped by the filtration system. On the other hand, the effect of combining an air filtration system with a moving robot has not been studied before. 
This paper shows that the mobility of robots augments air filtration to achieve well-covered protection of indoor people against scattered sources of airborne pathogen. 

As an overview, the contributions of this paper are:
\begin{itemize}
    \item Designed and built the Purdue Campus Patrol Robot -- a prototype for indoor mobile air filtration, which promotes effective, continuous and occupant-friendly protection in a classroom setting.
    \item Proposes a class of parameterized trajectory in a robot-friendly classroom arrangement, and optimizes the parameter to minimize the infection risk of people adjacent to the source, given the total number of people the robot is servicing.
\end{itemize}

The remainder of the paper is organized as follows: Section \ref{sec:odr} gives an overview of the urgency of the problem and the PCPR setup. The spreading dynamics of bioaerosol and the risks involved is further analyzed in Section \ref{sec:simple_single_source} with numerical simulation. Section \ref{sec:disinfection_sim} builds a generalizable model of virus spreading in classrooms, where the efficacy of our robot is analyzed and optimized. Section \ref{sec:exp} compares the optimization results with that obtained in an experiment using a mimicked source. Finally, Section \ref{sec:conclusion} draws the conclusion.
\section{Overview of the Purdue Campus Patrol Robot for Mobile Disinfection}
\label{sec:odr}

The PCPR (Purdue Campus Patrol Robot) 
integrates robotics into indoor public settings tasked to carry out repetitive and occupant-safe disinfection procedures. 
Due to the covert and contagious nature of the virus, symptom-less people still pose risk of spreading the virus to the susceptible, posing persistent pressure to sporadic outbreaks. Therefore, well-covered, efficient disinfection when people are gathering indoors is essential to suppressing community spread. 
Specifically, the classroom scenario serves as a simple but typical model for the types of disinfection during the \textit{reopening phase}. The people are statically located while obeying social distancing, while the transmission through both bioaerosol and contact of contaminated surfaces exist. The variety of transmission paths require a variety of complementary capabilities of periodical and effective disinfection during hours of operation. 

\subsection{Problem Statement}

Schools have been taking measures in arrangement to prevent community spreads during reopening. For example, Purdue University mandated seating spacing in classrooms of 2 meters longitudinally and 1.5 meters laterally. Some classrooms have additional fans set up at the doors for ventilation. However, the social distancing guidelines are based on either emperical guess or over-simplified assumptions, which may result to failure in protection \cite{Setti2020-6ftNotEnough}. As static air filtration and venting methods become an assistive measure to keep the indoor pathogen concentration low, they have coverage issues -- the source can be in any possible locations in the room, which means ideal airflow to vent or clean the pathogens is not guaranteed. On the other hand, placing an air filter in front of every occupant in the room is costly and not feasible when the number of occupants scale. Therefore, we are interested in how robots can suppress the spread by combining mobility with air filtration.

\begin{figure}
    \centering
    \includegraphics[width=0.45\textwidth]{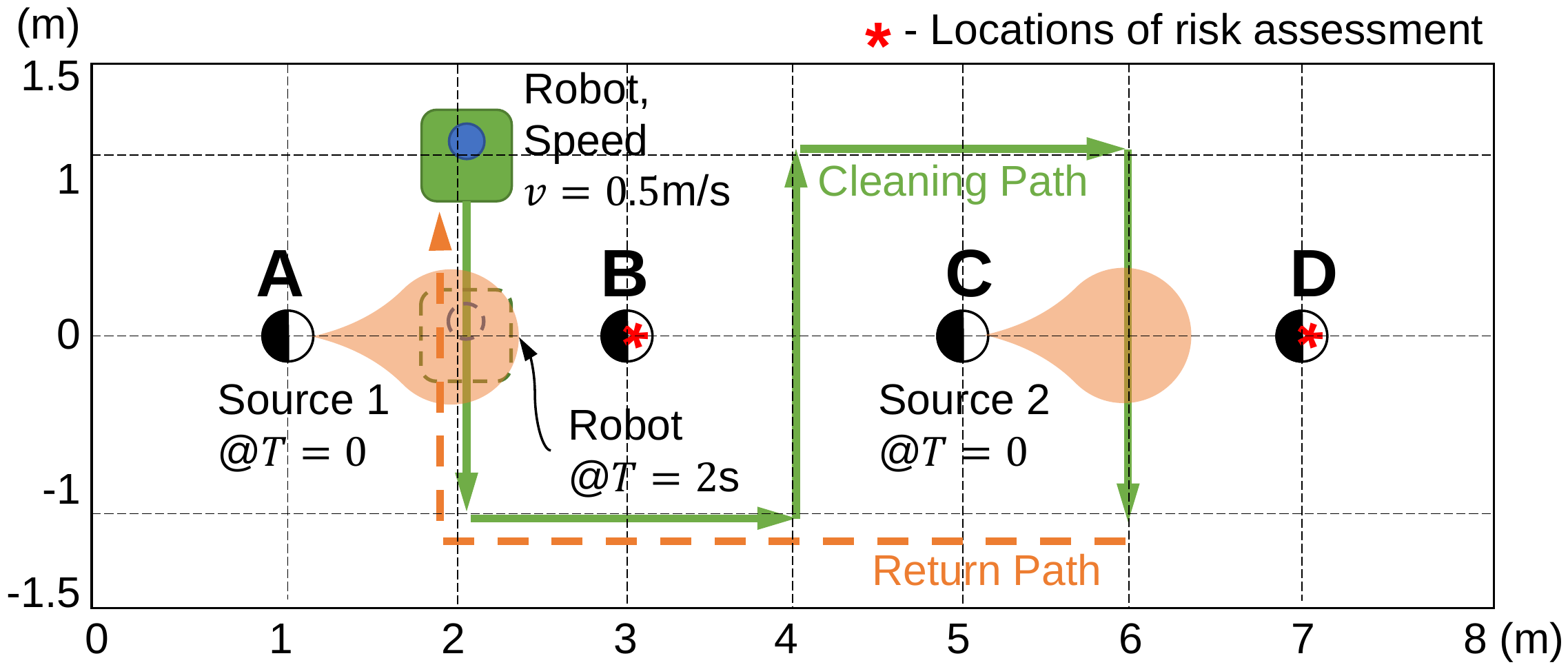}
    \caption{A simple trajectory of the robot when cleaning a space with four people}
    \label{fig:sim_rob_path}
\end{figure}

\begin{figure}
    \centering
    \begin{subfigure}{1.0\linewidth}
        \includegraphics[width = 1.0\linewidth, trim={5mm 35mm 12mm 35mm}, clip]{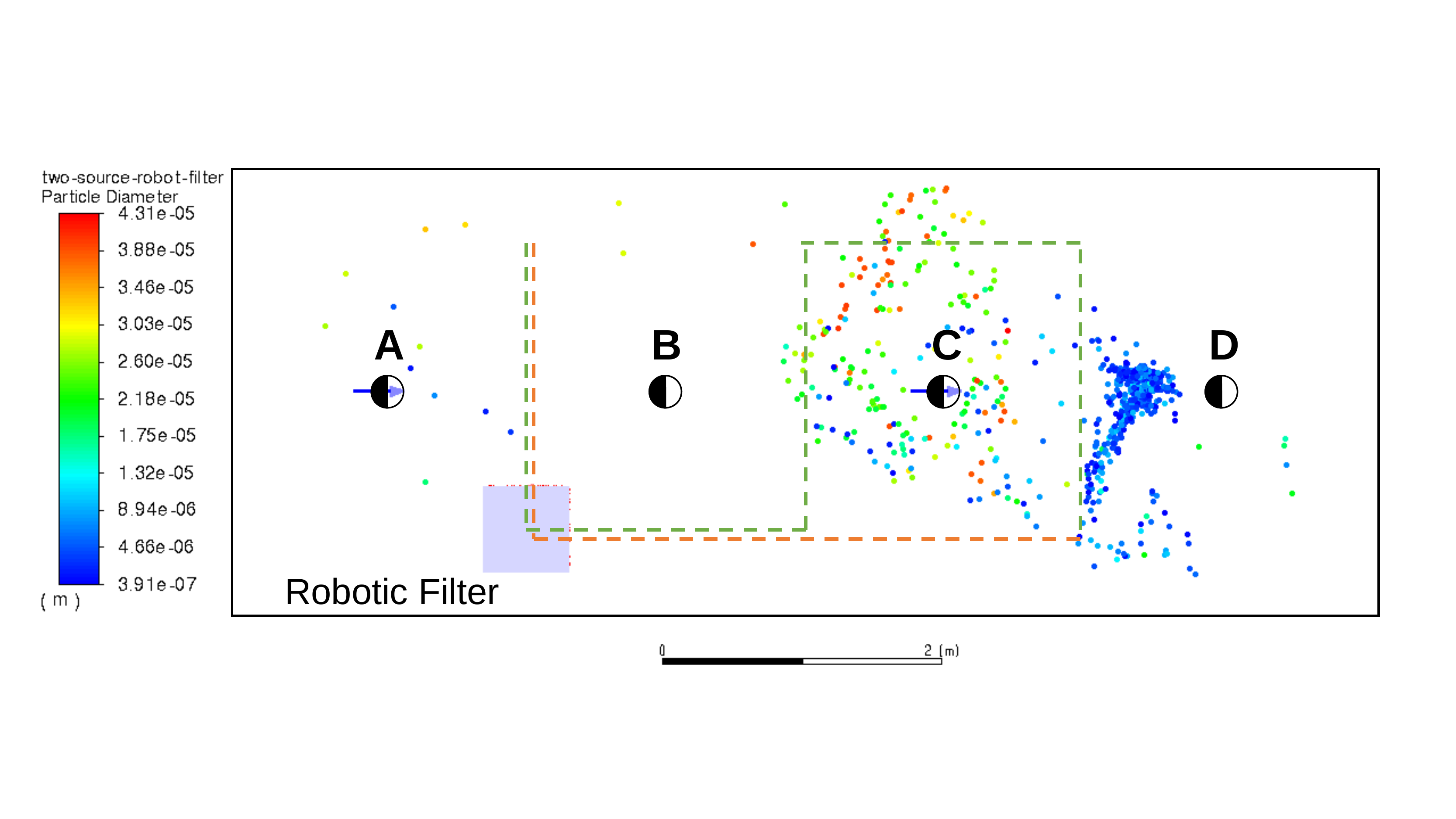}
        \caption{Robotic Filter Case}
    \end{subfigure}
    \begin{subfigure}{1.0\linewidth}
        \includegraphics[width=1.0\linewidth, trim={6mm 35mm 10mm 35mm}, clip]{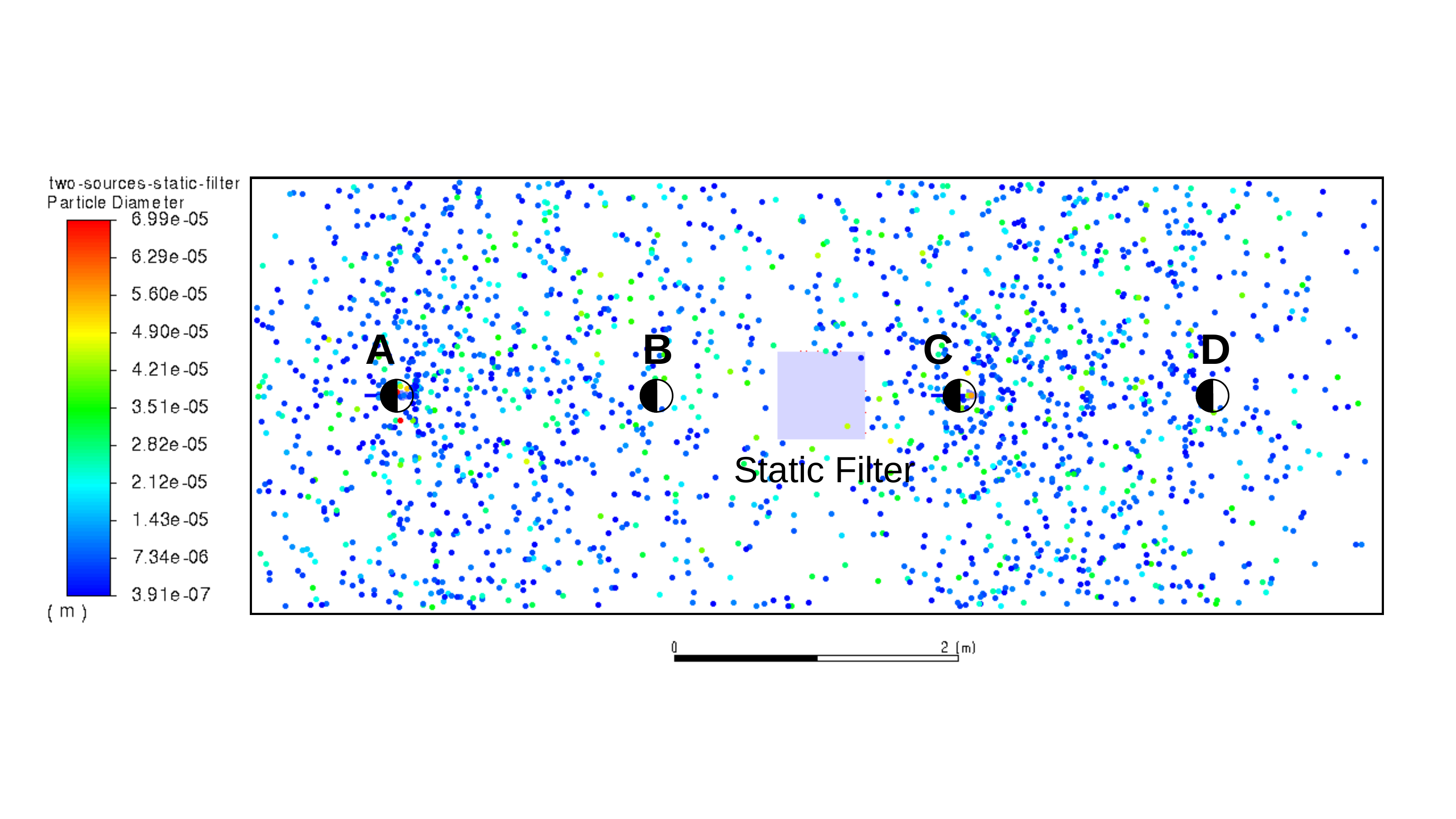}
        \caption{Static Filter Case}
    \end{subfigure}
    \caption{Saliva droplet distribution comparison at 1 minute after emission from two simulated coughing sources.}
    \label{fig:sim_robot}
\end{figure}

A simple comparative simulation is set up as Fig. \ref{fig:sim_rob_path} to illustrate the problem more intuitively: 4 people (A, B, C, and D) are placed in a room of 3\,m$\times$8\,m with coordinates as specified to follow social distancing guidelines, and all people faces to the right. In one case, a static filter is set at the geometric center to ensure least sum of distances to the four people, while in another case, the same filter is placed on a mobile robotic base that moves cyclically on the specified path across the middle points between people, with a specified velocity of $v=0.5$\,m/s. Since every person can possibly be the source, we specifically make persons A and C be the source of virus and they cough simultaneously. We evaluate the spreading pattern in the two cases using Computational Fluid Dynamics (CFD) simulation, which will be elaborated in Sec. \ref{sec:simple_single_source}. The distribution of particles after 1 minute, in both cases, are compared side by side in Fig. \ref{fig:sim_robot}, which clearly shows the advantage of the robotic filter over the static filter in cleaning speed and coverage. For persons B and D, slower cleaning of the plume means longer and faster accumulation of their dosages of the virus, hence poses higher risks of infection. The simple simulation signifies the improvements of a mobile robotic filter over a static filter, when multiple people are in the room without proper ventilation.








 
 

    

\subsection{Robot Composition}

The PCPR consists of three main parts: a holonomic mobile base, a spraying system, and an air filtration unit with internal UV disinfection. The omnidirectional holonomic base handles the navigation and obstacle avoidance with a planar LiDAR and two RealSense cameras. Its capability of holonomic motion eases constraints on path planning for directional payloads such as air filtration unit. Core safety techniques for collision avoidance are inbuilt to the base, ensuring the robot stops and circles around if it gets too close to any obstacle. With three parts combined, the robot can effectively perform autonomous multimodal disinfection on air and surfaces in an indoor space.

The remainder of this section describes the two forms of non-contact disinfection that the robot is capable of performing -- disinfectant spraying and mobile air filtration, among which mobile air filtration is novel. The spraying module has a static coverage radius, while the coverage dynamics of the air filtration system is not obvious. Hence, the analysis on the efficacy of air filtration will be expanded in Section \ref{sec:disinfection_sim}.

\subsubsection{Disinfecting Spray Module}

Spraying disinfectants is a hospital-proven measure of epidemic prevention \cite{otter2013role}. 
However, it is a labor-intensive job performed by staffs, 
who are exposed to two-fold hazard risks without extensive protection -- disinfectants usually cause irritation to human body, and the contaminated surfaces poses risks of infections. The spraying system on the robot uses a novel nanoparticle disinfectant solution and micro-spray system that can run either pre-scheduled or teleoperated, which removes staffs from the harsh working environment. 
The spray module has nozzles on both sides and rear of the robot, since spraying from the frontal side interferes with the perception sensors. The nozzles produce a cone-shaped micro-droplet cloud of disinfectant with an average droplets size of $\sim50\,\mu$m, covering a radius of up to 2.25\,m from the robot. An optional Electrostatic Spray Deposition (ESD) sprayer can be added to enhance coverage in corners and hard-to-reach areas.

\subsubsection{Novel Bernoulli Air Filtration Module}

Infectious bioaerosol droplets with diameters ranging from 0.1 to 10\,$\mu$m, which can suspend in air for hours, pose a major threat for horizontal transmission through inhalation.
The PCPR employs a novel air filtration module to increase the effectiveness of robotic filtration. The Bernoulli air filtration module brings medical-grade air treatment---a combination of particle filtration (a High Efficiency Particulate Air (HEPA) filter) and UV disinfection---to public indoor spaces, aiming to inactivate infectious bioaerosol droplets.


In our setup, the filtration efficiency, measured in percentage of the captured particles, is higher than 99.3\% for particles larger than 0.3\,$\mu$m, with capacity of 47\,L/s air. In the cylindrical air duct of the Bernoulli filter, two germicidal 2.5\,W UV lamps with surfaces irradiation of 5\,mW/cm$^2$ are adopted. Being contained inside the cylinder, the UV light poses no exposure risk to passersby.


The flow tract design of the Bernoulli module, i.e. the AeroMINE\texttrademark (Motionless, INtegrated Extraction) technology, originates from wind energy applications \cite{houchens2019novel}. Using a pair of mirrored perforated airfoils and a perforated cylindrical tower, 
a low pressure zone is created between the airfoils which helps to drive the air from surroundings into the intake of the system, which then will be driven by an axial fan in the cylinder through an filter. This design achieves large intake area without using a large-diameter fan, thus improves energy efficiency.  
The large intake area creates a large swept volume to trap the pathogen as the robot moves, which is ideal for robotic disinfection.  

\section{Transmission Pattern of Airborne Pathogen and Dosage Assessment}
\label{sec:simple_single_source}


To investigate the transport and dispersion of saliva droplets in an indoor environment and cast insight on how robots can help, CFD simulations of a single coughing source were performed. Based on which, we extracted the evaluation metrics to differentiate the robotic filter versus a conventional stationary filter, and created snapshots for estimating the effects of a robotic filter.

\subsection{CFD Simulation Setup for Single Coughing Source}\label{sec:cough_snz_model}

Respiratory pathogens can be emitted in the form of bioaerosol when an infected person sneeze, cough, or even breathe. Specifically, we are interested in the coughing case since it is more covert and frequent compared to sneezing, while having similar emission magnitude (in terms of jet velocity) in some conditions \cite{Tang2013SneezeCoughVel}. We also consider a worst-case scenario when no facial coverings is present, as people tend to relax caution on the rules over time. The results from the CFD simulation provide building blocks for optimizations afterwards, and to show that 2\,m (6\,ft) of social distancing is not always sufficient.

With ANSYS Fluent 2019, the room geometry is built as a rectangular domain of 3\,m $\times$ 8\,m $\times$ 3.5\,m (width $\times$ length $\times$ height). 
For simulating the turbulent airflow of coughing cases, the Reynolds Average Navier-Stokes (RANS) with the Renormalization Group (RNG) $k$-$\varepsilon$ turbulence model was used in the simulations for its accuracy, computing efficiency, and robustness \cite{hang2014influence}. 
The Lagrangian method was used to track saliva droplets with hybrid implicit and trapezoidal schemes. To emulate a real human cough, saliva droplets are ejected from the mouth with velocity profiles based on previously performed measurements \cite{gupta2009flow, aliabadi2010cfd}. In our case, 14000 droplets with diameters ranging from $2r_P$ = 1\,$\mu$m to 500\,$\mu$m and following Rosin-Rammler distribution \cite{Lindsley2012CoughParticle}, were injected from the inlet for the coughing event. The coughing jet lasts 0.61\,s with a peak velocity around 22.06\,m/s occurring at 0.066\,s. 

\begin{figure}
    \centering
    \includegraphics[width=1.0\linewidth]{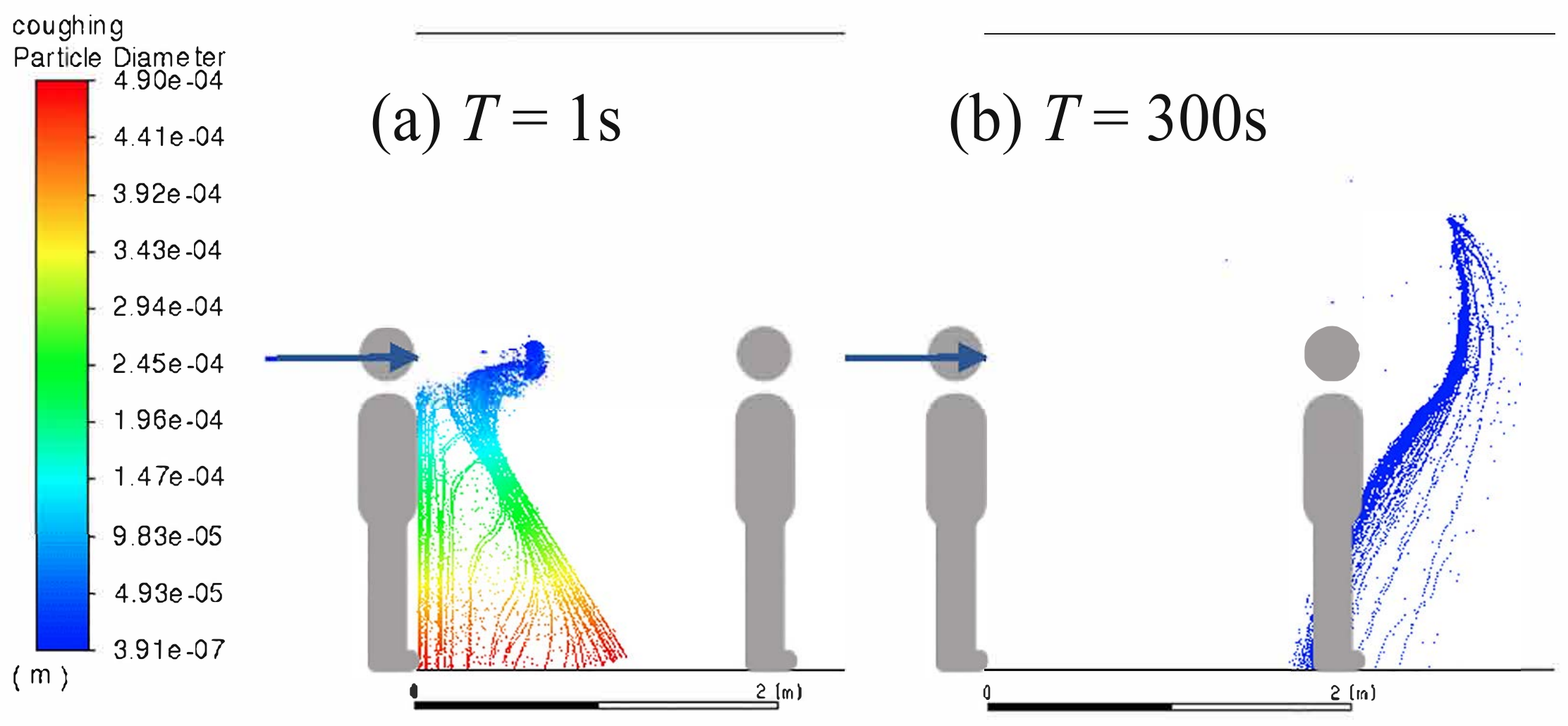}
    \caption{Saliva droplet spreading from a coughing source (blue arrow) at (a) $T=1$s and (b) $T=300$s after the emission. The horizontal scale denotes 2\,m social distancing.}
    \label{fig:IEEE_cough_sneez}
\end{figure}



The total time of analysis is 5 minutes with non-uniform time steps for both discrete (droplets) and continuous phases (air), to show that the particles can spread over 2\,m of distance, as shown in Fig. \ref{fig:IEEE_cough_sneez}. Since the velocity has been dissipated, the infectious plume will cause continuous exposure to the adjacent people. These results reveal the need for air filtration that can effectively capture the droplets before dispersion.

\subsection{Dosage Criteria}

Realistic data is used to evaluate the risk of a person during the spreading process of the saliva plume of pathogen. Since the saliva is diluted in the air, the virus concentration, measured in PFU/m$^3$ of air, is averaged in a grid-wise manner in the horizontal plane because the particles $P_i$ are discrete objects. Since the robot and people move in a 2-D plane, the 3-D virus concentration can be further averaged vertically for evaluation. The averaging range is the entire room height $H$, in case of possible vertical perturbation of the air. Denote the concentration of virus in saliva as $c$, the averaging grid size as $2l$, the concentration $C$ of virus in the air at time $t$ at grid centered $(x,y)$ is thus generated from the particle distribution $P$ as:
\begin{equation}
    C_P(x,y,t) = \frac{kc}{4Hl^2}\cdot \frac{4\pi}{3} \sum_{P_i(t)\in\{(x-l, x+l)\times(y-l, y+l)\}} r_{P_i}^3,
\end{equation}
where $k$ is the scaling value due to liquid evaporation (hence virus consolidation) in the droplets, and $r_P$ denotes the radius of the particles. Here, we set $c=10^6\, \textnormal{PFU/mL}$ \cite{Schijven2020SARS-CoV2-Concentration} and $k=10$ \cite{aliabadi2010cfd} in the analysis followed.

Since a normal human has a tidal volume of about $V=0.5$L and breathing rate of 12 per minute at rest. Because the tidal volume equivalents to that of a sphere with diameter of 0.1\,m, we set the grid size to be $2l=0.1$m. The ``capture grid" is defined as a grid where an observer stands. The dosage per breathing of the observer is derived from the average particle concentration on the capture grid over 5 seconds. Hence the total dosage the observer receives through an exposure period of $t_e$ (seconds) is obtained:
\begin{equation}
    D_P(x,y)_\textnormal{obs} = \sum_{i=0}^{t_e/5-1}\int_{0}^{5}\frac{ C_P(x,y,5i+t) V}{5}\textnormal{d}t.
    \label{eqn:dosage}
\end{equation}

To increase the robustness of the dosage estimation with respect to location error of the observer, a max dosage $\hat{D}_P(x,y)_\textnormal{obs}$ is taken as the maximum dosage on the 3-by-3 grids centered at $(x,y)$. 
As a reference for severity of exposure, studies of SARS-CoV indicates an at most 5\% chance of developing illness from the virus when exposed under virus dosage of 10\,PFU \cite{Watanabe2010SARS-infectivity}. 
Under this criteria, we will build a generalizable model on mobile air filtration in classroom setting in Section \ref{sec:disinfection_sim}.

\section{Predictive Assessments of Classroom Social Distancing \& Robotic Filtration} 
\label{sec:disinfection_sim}


Based on the CFD simulation results, this section assess the risk of virus spreading to the adjacent people. Specifically, the classroom seating model and periodic cleaning model are built, providing parameterized design and optimization approach to efficient robotic air filtration servicing specified number of people.



\subsection{Modeling of Transmission and Robotic Cleaning of Airborne Pathogen in a Classroom Setting}

\begin{figure}
    \centering
    \includegraphics[width=0.9\linewidth]{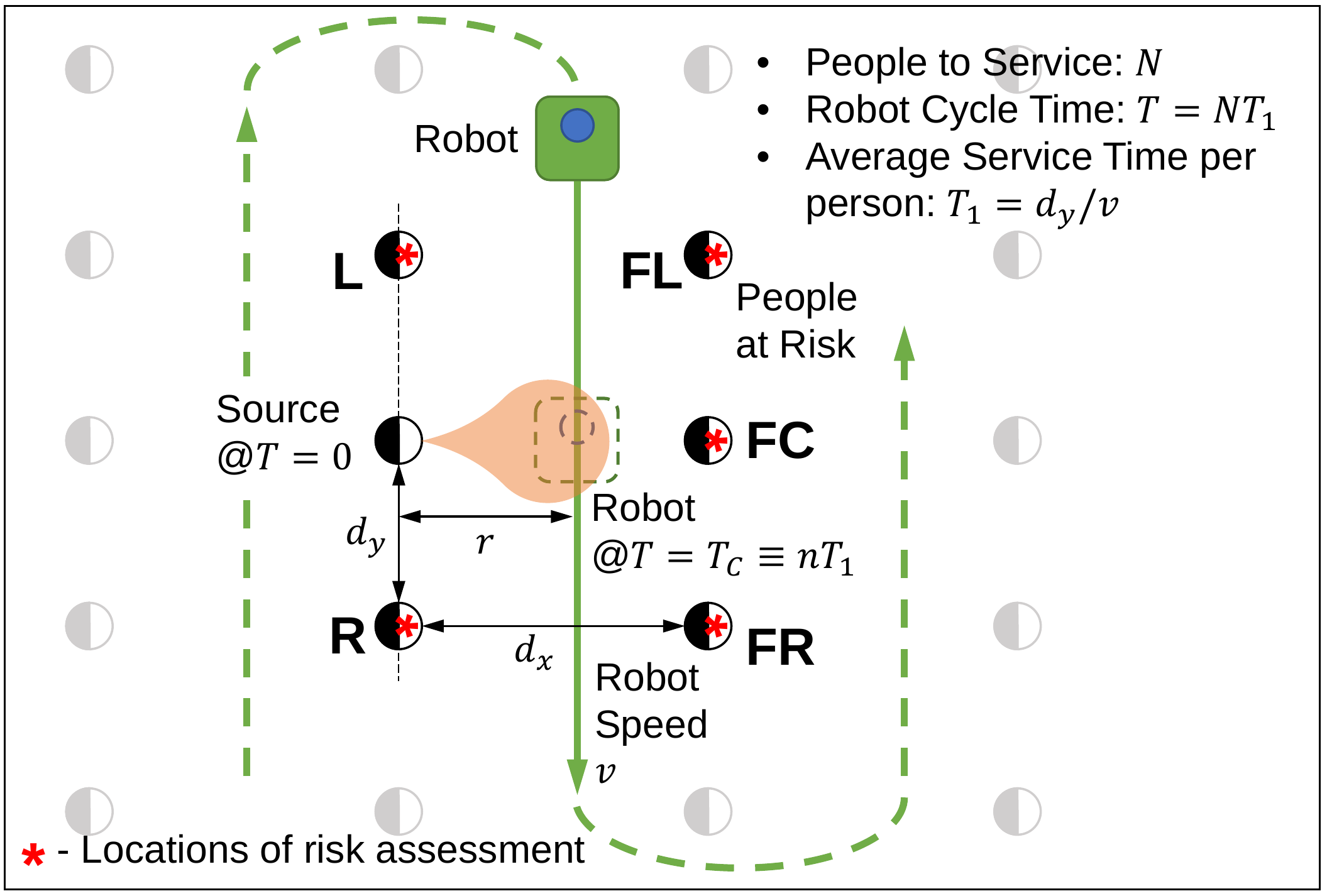}
    \caption{Geometric parameters of the indoor spread and cleaning model. Observers are the people at risk with red asterisks.}
    \label{fig:robot_classroom_cleaning}
\end{figure}

Without losing commonality, occupants (students) are assumed to be static, sitting in a grid with specified and fixed spacing $(d_x, d_y)$, and facing the same ($+d_x$, or right as seen in the figure) direction. As a result, we assume the plume of particles from an emission event does not spread backwards. The virus carrier is unaware of becoming the source of spread, hence the exact source location is unknown. However, when an emission does occur, we identify the five people seated to the front and sides of the source to be the most vulnerable, based on the spreading pattern of the plume as shown in Section \ref{sec:simple_single_source}. These people are denoted as ``observers", whose infection risks of are evaluated with virus dosage as \eqref{eqn:dosage}.

Since robotic filtration works by proactively moving to and cleaning up the plume before it spreads, the robot needs to pass by every occupant in the room to achieve the cleaning task. The reason for cleaning the space near every individual is two folds: only humans can emit the infectious plume, and the virus carrier is randomly hidden among the occupants. For occupants seated in a grid pattern, the simplest pattern of robot motion that 
traverses each occupant is sequentially passing by the front of each row (from left to right of one person, or vice versa) at distance $r\in(0,d_x)$, in a straight line and uniform speed $v$, before turning around to clean the next row, as shown in Fig. \ref{fig:robot_classroom_cleaning}. Compared to moving in column direction, row-directional motion enables the robot to pass in front of each occupant, where the probable concentration of particles is the maximum. We neglect the time the robot spend to switch rows or return after a cycle, because they are relatively fixed depending on the room layout, hence can be added afterwards as overhead. To summarize, the cycle time of the robot, $\tau$ is bounded by the total number of people $N$ the robot is passing:
\begin{equation}
    \tau \equiv NT_1 = Nd_y/v
\end{equation}
where $T_1$ is the service time for each person, further bounded by the lateral spacing of the occupants $d_y$ and the velocity of the robot. We further assume that the robot has a quantum release time, such that if the source emits at $T=0$, the possible times that the robot happens to be in front of the source are $T_C = nT_1, \ \ \ n=0,1,\cdots, N-1.$
Since the robot cleans the room periodically, passing $N$ people has the worst-case latency of $T_{C,\max}=(N-1)T_1$ after the emission from the source.

\subsection{Optimization of Path Parameters}

The efficacy of the air filtration is defined as the reduction ratio of maximum cumulative dosage per emission event among the five observers, compared to the case without any filtration (i.e. natural precipitation and natural diffusion). Since the exact time of emission is unknown, the objective is set to minimize the maximum dosage for all cleaning delays $T_C$, given a fixed $N$. The parameterized path of the robot left us with two variables to tune -- robot speed $v$ and distance $r$ from each row of people. 
Therefore, the optimization formula can be expressed as:
\begin{equation}
\begin{array}{l}
    \{v_N^*, r_N^*\} = \underset{v\in(0,v_{\max}), r\in(0,d_x)}{\arg\min} \max\left(\hat{D}_{P_n}(x,y)_{\textnormal{obs}}\right): \\
    (x,y)\in\{(0,\pm d_y), (d_x, 0), (d_x, \pm d_y)\}; P_n\vert_{\forall T_C}
\end{array}
\end{equation}



Since CFD simulation of exact particle distribution is extremely computationally-intense and time-consuming, it is impossible to perform iterative computation for path optimization of the robotic filter. Therefore, a low-order approximation is developed for the evaluation of the optimization objective. As a preliminary assumption for path parameter optimization, we assume the spread of the plume is not macroscopically affected by the motion of the robot given that the robot is moving slowly. 
From Section III-A we identified that the initial momentum of the particles dissipated quickly after emission, therefore the spread can be approximated with a decayed diffusion to extrapolate the CFD data with step size $\Delta t$:
\begin{equation}
    C(x,y,t+\Delta t) = k_t^{\Delta t}\left(C(x,y,t) + k_d \Delta t \cdot \nabla^2 C(x,y,t)\right),
    \label{eqn:diffuse}
\end{equation}
where $C$ is the field to be diffused. $k_d$ is the equivalent diffusivity, and $k_t$ is the time constant of the exponential decay, both estimated using distribution of virus concentration from the CFD results in the time range from 30 to 60 seconds. The estimated values are $k_d \approx 0.003$\,m$^2$/s and $k_t \approx 0.98$, but the actual values depend on the air flow of the room.

During each dosage accumulation time step, the effect of the robotic air filter is simplified by decreasing a factor of $k_f=Q\Delta t/V_r$
on the virus concentration and influx at each discretized grid that is within the footprint of the robot. Here $Q$ is the flow rate of the filter, and $V_r$ is the volume of the column formed by the robot footprint. The decrements are also decayed and diffused using \eqref{eqn:diffuse} before propagating to the next dosage accumulation time step.


The dosage evaluation ranges from 0 to 600\,s for convergence, and uses a step size of 1\,s. The evaluation is implemented in MATLAB with the optimization process realized through the \texttt{fmincon} function. CFD result from a single source is referred to in the time range of 0 to 60\,s, as the velocity of the plume after 60\,s is small and therefore susceptible to perturbations. 
The optimization results show that the efficacy of the robot, with a given $N$, increases with $r$ and $v$. The efficacy of the robot decreases monotonically with the increase of cleaning cycle $\tau$, i.e. the number of people $N$ it is passing. However, with large passing distance the cleaning delay $T_C$ becomes less sensitive. The optimization result also show that among the five observers, the one in the Front Center position is the most vulnerable, in that it receives the highest dosage.

\begin{figure}
    \centering
    \begin{subfigure}{0.495\linewidth}
    \includegraphics[width=1.0\linewidth, trim = {0 0 0 5mm}]{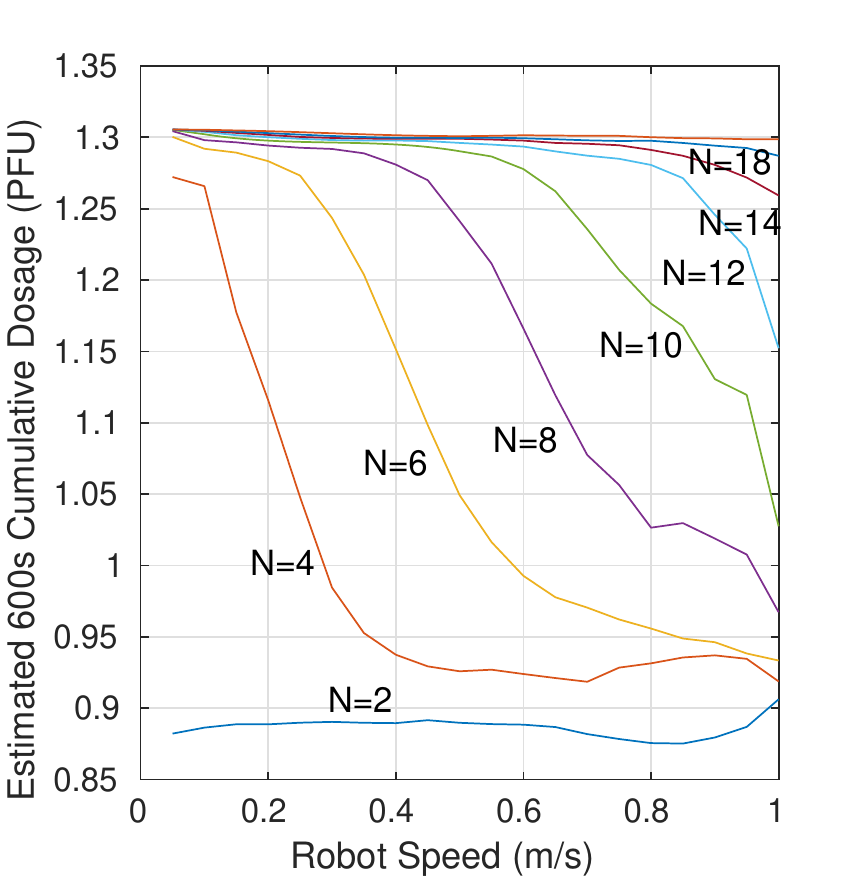}
    \caption{Dosage vs. Speed, \\assume passing at $r=1.3$\,m.}
    \end{subfigure}
    \begin{subfigure}{0.475\linewidth}
    \includegraphics[width=1.0\linewidth, trim = {0 0 0 5mm}]{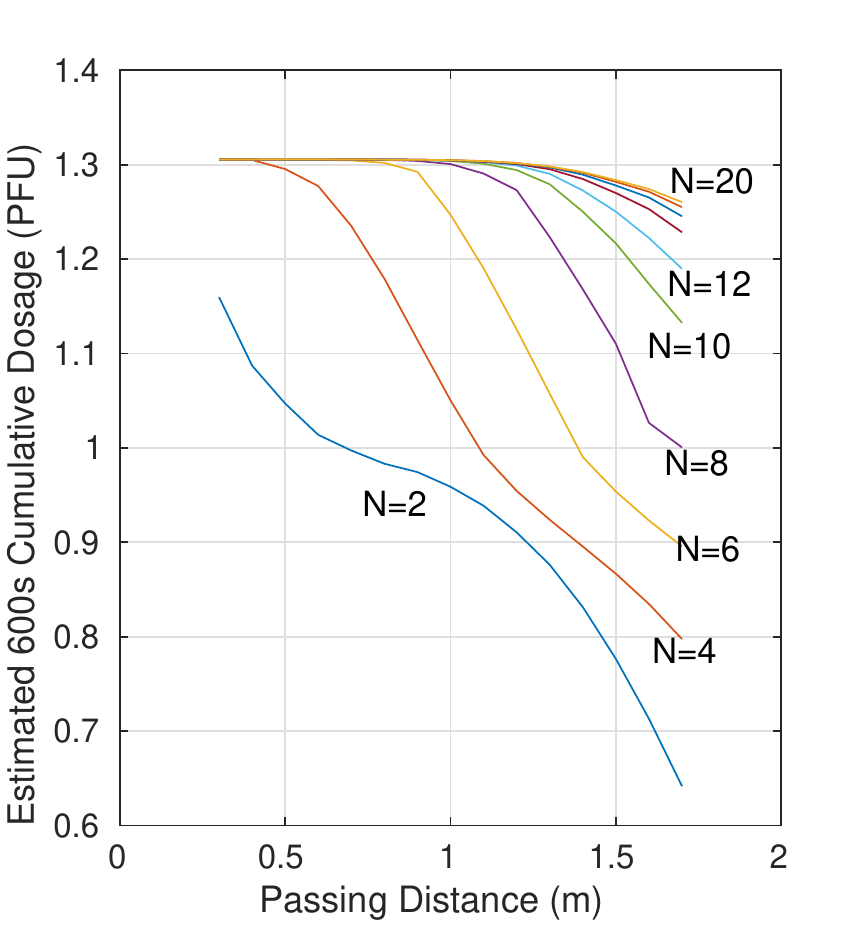}
    \caption{Dosage vs. Passing Distance, \\assume passing at $v=0.5$\,m/s.}
    \end{subfigure}
    \caption{Projected trend of cumulative dosage with a robotic filter during a single cough event, at Front Center point, with respect to design parameters and people serviced.}
    \label{fig:slices}
\end{figure}

The projected cumulative dosage per emission for the Front Center position, under different passing distance and velocities, is shown in Fig. \ref{fig:slices}. An experiment is carried out to validate the optimization result in real-world indoor airflow, and using a mimicked source.

\section{Experimental Validation on the Efficacy of Robotic Mobile Air Filtration}
\label{sec:exp}




Although optimizations in \ref{sec:disinfection_sim} showed promising results that the robot is able to reduce the dosage received by the people closest to the source, the real-world airflow may affect the results drastically. As a validation, we carried out an experiment, physically duplicating the case in Fig. \ref{fig:robot_classroom_cleaning} in a realistic indoor environment.

\begin{figure}
    \centering
    \includegraphics[width=0.81\linewidth]{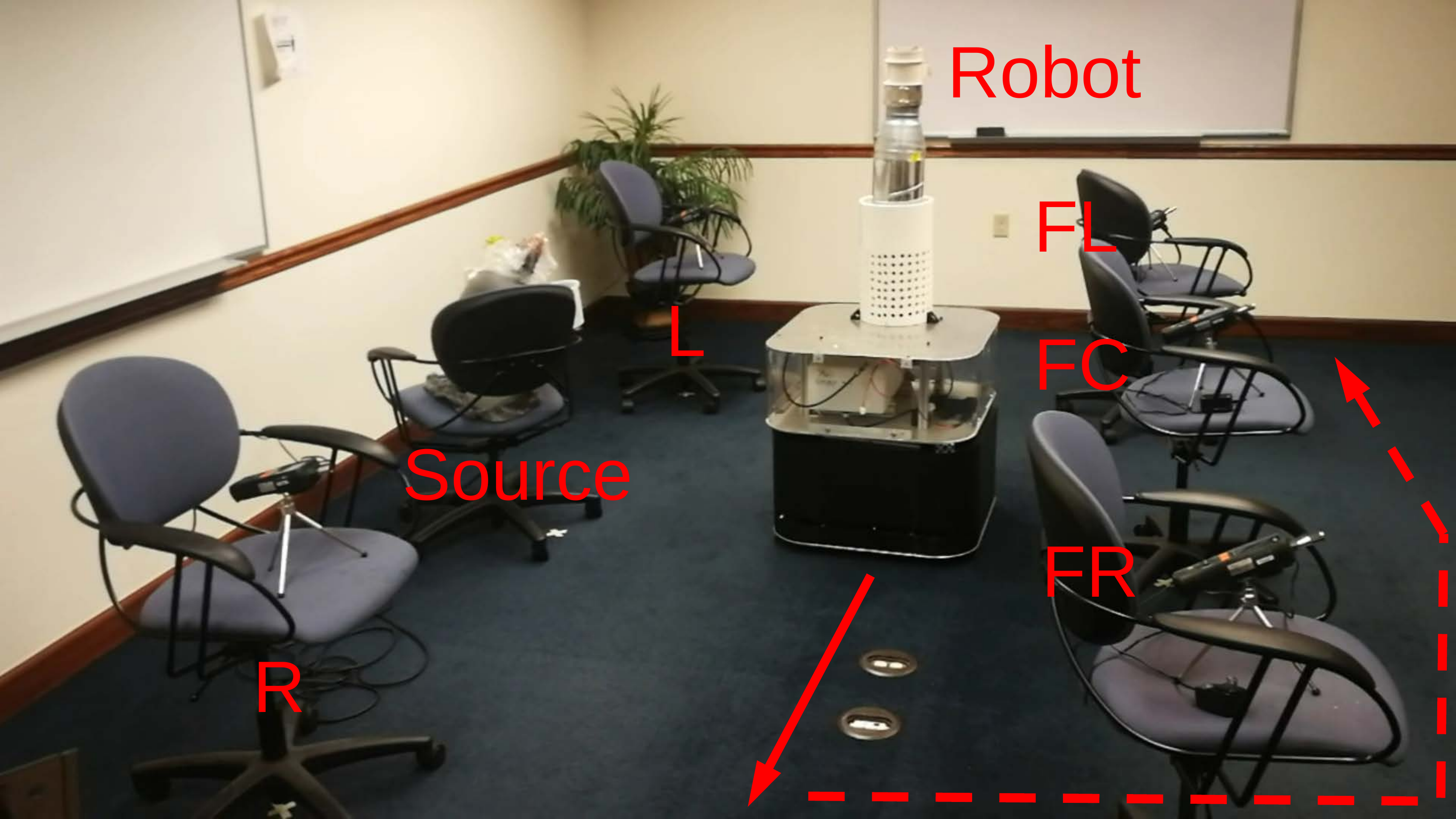}
    \caption{Experiment with PCPR and a mimicked source. The airfoils of the Bernoulli Air Filter were removed to match the simulation and optimization assumptions. The red arrows indicate the robot's cleaning and return paths.}
    \label{fig:expSetup}
\end{figure}
As shown in Fig. \ref{fig:expSetup}, we align chairs in a room of 5\,m$\times$6\,m, following social distancing guidelines at Purdue, with $d_x=2.0$\,m and $d_y=1.5$\,m. We assume the center seat at the rear is the source, and five article counters (ExTech VPC300) are placed at the observer (Front Left, Front Center, Front Right, Left, and Right) locations. Since the magnitude of emission velocity affects spreading speed of the plume, hence affects the cleaning efficacy of the robot, the horizontal velocity of the emission should match that of a normal human cough as discussed in Section \ref{sec:simple_single_source}. The number of particles emitted, however, can be scaled up to obtain better signal to noise ratio over the background particle counts.

In the experiment we use a vector fogger with pitch-up angle of 65$^{\circ}$ and copper mesh at the nozzle to obtain a peak horizontal speed of 22\,m/s. 
A plastic bag loosely covers around the nozzle and the mesh to collect the backflow and the droplets within. The source was turned on for 2\,s each, and the measurements at the five observers last for 10 minutes. The particle counter reads the accumulative counts for particles of diameters 0.3, 0.5, 1.0, 2.5, 5.0, and 10\,$\mu$m, hence the total 10-minute dosage based on volume of inhaled particles can be calculated with \eqref{eqn:dosage}. The background is measured and subtracted prior to the start of each cumulative measurement series. Since the number of particles is scaled up compared to a normal human cough, we compare the relative dosage by scaling the highest value observed down to 1.

Following the guidelines obtained from Section \ref{sec:disinfection_sim}, we set the passing distance of the robot to be $r=1.3$\,m, which is the maximum distance without the robot hitting the legs of the chairs in the front row. The speed of the robot moving through the cleaning zone (i.e. between the two rows) is set to be $v=0.5$\,m/s. After completing each pass, the robot moves out of the cleaning zone and circulates back to the starting point within specified time. The circulation behavior simulates the cleaning of other spaces, with the cycle time set to $\tau=30$s, $36$s, and $42$s, corresponding to $N=10, 12, 14$, respectively. The source is released such that the cleaning delay $T_C = \tau-3$,  the maximum cleaning delay possible. Three additional cases are set up as comparisons: A) ``Near Static Filter": the static filter is placed at 1.3\,m in front of the source; B) ``Far Static Filter": the static filter is placed 2\,m in front of the Front Center person, simulating a centralized filter placed at the center of a larger room; C) no air filtration is provided.

The normalized dosages at the five observers in different cases are shown in Fig. \ref{fig:dosage_compare}. From the results we conclude that our robot setup is able to decrease the dosage that people around the emission source receive by at least 20\% when servicing 14 people, which is the capacity of an average classroom when social distancing guidelines are observed. The cumulative dosage slightly decreased as $N$ goes from 10 to 14, which as we interpret, is due to increased background that drowns out the tail of the dosage accumulation, as these experiments happens in chronological order. The experimental result outperforms that obtained from the optimization simulation, and the dosage distribution among the five observers are not always symmetric, which we discuss as follows:

\begin{figure}
    \centering
    \includegraphics[width = 0.95\linewidth]{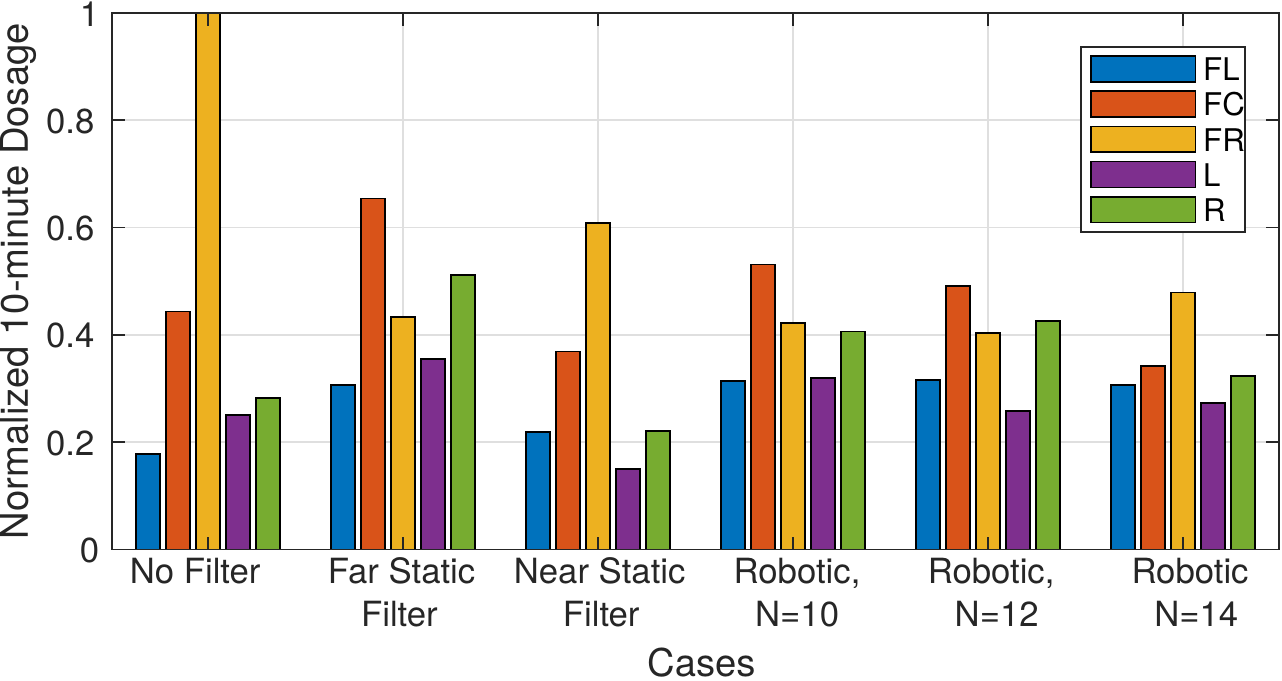}
    \caption{Cumulative dosage comparison of the five assessment locations under different filtration settings.}
    \label{fig:dosage_compare}
\end{figure}

\subsubsection{Wake of the robot attracts particles to boost cleaning effects}
The motion of the robot creates a low-pressure zone behind the robot that traps particles along with the robot. The effect was seen through CFD simulation as well. The wake, as a bonus from mobility, improves cleaning efficacy while also poses design considerations on the shape of the robot. The airfoils of the Bernoulli filter is an example that utilizes the wake to keep stable traction to the particles as the robot moves.

\subsubsection{Volatility of indoor airflow}
Some of the experiment results show biased dosage distribution to the right, which is due to the design of air conditioning system in the room. The volatility of indoor airflow also demonstrated the robustness of the robotic filter, compared to the Near Static Filter case, in cleaning particles that deviated from their theoretical positions.

\subsubsection{Considerations on classroom arrangements} In order to utilize the $N$ most effectively, for robotic filtration, classrooms should be arranged in the lateral direction to reduce row switching time for the robot.

\subsubsection{Real-time environment assessment for pathogens to optimize cleaning efficiency} PCPR seeks to integrate biosensors for pathogen detection, which in real time assess the robot surroundings. The real-time environment information will be invaluable to dynamic planning and optimizing the efficiency of air and surface disinfection,
\section{Conclusions}
\label{sec:conclusion}

The Purdue Campus Patrol Robot combines mobility in robots with air filtration, which achieves effective, robust, and occupant-friendly cleaning of classrooms. The robot, capable of serving more than 10 students at once, effectively reduces worst-case dosage that a people can receive in case of a cough in a classroom. As a result, the safety of students can be further enhanced while sporadic outbreak further prevented during school reopening from COVID-19. The workflow of modeling and optimization provides insights to future design of the robot and arrangement of robot-friendly classrooms.

\section{Acknowledgements}
This work was funded in part by Intel Corp COVID-19 fund, the Indiana Manufacturing Institute, the NSF RoSe-HuB Center, and NSF under CNS-1439717.

\bibliographystyle{IEEEtran}
\balance
\bibliography{references}
\end{document}